# Background Bias Assessment and Mitigation in Plant Disease Detection Using EfficientNet-B0 with Image Processing

Majed Molhi [1] *
*Department of Mathematics, University Institute of Sciences*
*Chandigarh University*
Mohali-140413, Punjab, India
majedmolhi@gmail.com

Nishi Gupta [2]
*Department of Mathematics, University Institute of Sciences*
*Chandigarh University*
Mohali-140413, Punjab, India
nishi.gupta.phd81@gmail.com

Ali Algumaei [3]
*Faculty of Information Science and Technology*
*Universiti Kebangsaan Malaysia (UKM)*, Malaysia
aliderhemqaid@gmail.com

***Abstract*— Plant disease detection using deep learning has achieved remarkable accuracies exceeding 99% on benchmark datasets. However, these metrics may mask a critical vulnerability: models may be taking advantage of false background correlations instead of learning real pathological features. This study presents a systematic investigation of background bias in plant disease detection through a two-stage training framework with quantitative validation. We propose a three-condition evaluation protocol testing models on original, leaf-only, and background-only images to directly measure background dependency. Our approach combines standard transfer learning using EfficientNet-B0 and manipulation of background information during optimization. Experimental results on the PlantVillage dataset (20,638 images, 15 disease classes) show that while our model achieves 98.97% accuracy on the original images and maintains 96.62% accuracy on leaf-only images, the performance drops to 23.58% on background-only images. The calculated background bias (BBC) coefficient of 0.762 indicates that a large part of the model's predictive performance depends on leaf-related features rather than background distortions. This research offers clear evidence that even with the most advanced models, environmental shortcuts can be leveraged, and this highlights the importance of bias evaluation. The approach that has been used has laid a foundation for the development of agricultural AI that is ready for deployment. The implementation, models, and evaluation process are all publicly available.**

## I. Introduction

Plant diseases significantly impact global food security, making early and accurate detection essential for effective intervention (Savary et al., 2019) [1] . While manual inspection methods are still ubiquitous in many agricultural settings, these approaches rely heavily on subjective experience and expertise and in large farming operations are simply unmanageable. In many cases, the unavailability of trained plant pathologists poses a considerable barrier to timely disease management.

Recent advances in deep learning, particularly CNNs with ImageNet-based transfer learning, have achieved near-perfect accuracy on controlled benchmarks such as PlantVillage.

Despite these advances, an important vulnerability remains largely unexamined: models are often evaluated solely by accuracy, without validating whether predictions rely on genuine pathological features or spurious background correlations. This concern is amplified by the widespread use of PlantVillage, where images are captured under highly controlled background conditions [2]. If image capture conditions are systematically associated with disease classes—for example, healthy leaves are consistently imaged against one type of background while diseased leaves appear against another—models can achieve high accuracy by learning environmental patterns rather than disease symptoms.

In this work, we focus specifically on contextual background bias, where visual cues unrelated to plant pathology—such as capture setup, illumination, or background texture—become correlated with disease labels. This form of bias is distinct from sensor-level noise or low-level acquisition artifacts, and directly impacts model robustness when deployed in real agricultural environments with unconstrained backgrounds. Such background-dependent models may perform well in laboratory benchmarks yet fail under real-world agricultural conditions.

The work compensates for the identified shortage by deploying a dual-phase training protocol for reducing background bias without sacrificing high-precision levels. Three main contributions emerge as follows: a three-condition evaluation protocol (original, leaf-only, background-only) with the Background Bias Coefficient (BBC) as a metric of background dependence; (2) a two-stage training approach that combines transfer learning with targeted background manipulation to enforce learning of disease-specific features; and (3) the first empirical quantification of background bias in the application area of plant disease recognition.

## II. RELATED WORK

### A. *Introduction*

Deep learning-based computer vision approaches have achieved reported accuracies exceeding 99% in plant disease detection. However, a critical limitation remains largely overlooked: models may rely on spurious background correlations rather than genuine pathological features. This raises concerns regarding the reliability of published results for real-world agricultural deployment.

### B. *Early Deep Learning Approaches*

Early studies transitioned plant disease detection from handcrafted features to deep learning. Mohanty et al. (2016) [3] achieved 99.35% accuracy on PlantVillage using ResNet50, VGG16, and GoogleNet, demonstrating the power of deep features, but without assessing whether predictions relied on disease characteristics or background cues.

Subsequent studies reported similarly high performance while keeping the same focus on evaluation. Lakshmanarao and Supriya (2022) [4] attained an accuracy of up to 99% utilizing VGG16, ResNet50, and Inception-V3, whereas Nguyen et al. (2022) [5] integrated K-means segmentation with VGG-19 to get an accuracy of 99.72% on tomato leaf pictures. However, it remains unclear whether these gains reflect reduced background dependence or simple computational optimization.

### C. *Advanced Techniques*

#### 1) *Transfer Learning and Efficient Architectures*

Transfer learning has become the most common way to get high accuracy with small agricultural datasets. The EfficientNet family has shown strong potential. Ali et al. (2025) [6] found that fine-tuning EfficientNet-B0 got 99.69% accuracy, which is a good balance between performance and computational efficiency. Lightweight architectures like MobileNet have also become popular in situations where resources are limited. Fathima and Booba got 96% accuracy by combining MobileNet with Local Binary Pattern (LBP) features. Despite improved efficiency, these approaches largely retain the same evaluation limitation by focusing on accuracy without verifying whether models learn disease-specific features.

#### 2) *GANs and Attention Mechanisms*

Recent studies explored GAN-based augmentation and attention mechanisms to address data imbalance and feature localization. Seshaiah and Vandana (2024) [7] utilized GAN-based augmentation in conjunction with SRCNN, attaining an accuracy of 95.69%; nevertheless, this augmentation may inadvertently perpetuate existing background biases. Karthikeyan et al. (2025) [8] developed an attention-based CNN-SEEIB, achieving 99.79% accuracy with little inference delay. Although attention systems seek to highlight prominent regions, the lack of explicit confirmation renders it ambiguous whether they prefer diseased traits or contextual cues.

#### 3) *Feature Extraction and Explainability*

Hybrid methods that combine deep learning with traditional feature extraction have shown promise. For example, SVM classifiers that use features from CNNs have high accuracy rates, which suggests that the quality of the features is often more important than the choice of classifier. Explainability efforts, particularly Grad-CAM, have provided qualitative insights into model attention. For instance, Tamim et al. (2025) [9] integrated Grad-CAM into the InsightNet framework to visualize decision regions. However, such methods remain largely qualitative and do not explicitly quantify the influence of background features on model predictions.

Many studies have explored the "Clever Hans" effect and shortcut learning in vision models, frequently utilizing explainability techniques like Grad-CAM to visually examine attention regions [10][11]. Although these methods yield qualitative insights, they lack quantitative validation regarding the capability of background features alone to facilitate classification. The proposed three-condition evaluation procedure directly quantifies prediction performance through controlled feature isolation, facilitating explicit numerical evaluation of background dependency.

### D. *Background Bias Gap*

Despite achieving high reported accuracies, existing studies rarely verify whether models learn disease-specific features or exploit background correlations. The PlantVillage dataset—used in 67 of 160 surveyed studies (Pacal et al., 2024) [2]—consists of laboratory-acquired images with controlled backgrounds that may induce systematic bias. When capture conditions are related to disease labels, models can learn patterns in the environment instead of disease characteristics, which can lead to very high accuracy. This problem hasn't been looked at very much, which makes it hard to trust real-world use where backgrounds are always changing.

### E. *Summary*

The plant disease detection literature is largely characterized by an emphasis on maximizing accuracy, strong dependence on controlled laboratory datasets, and the absence of systematic bias evaluation frameworks capable of distinguishing pathological features from environmental artifacts. Although recent architectural and methodological advances continue to improve benchmark performance, they remain confined to evaluation settings that do not assess feature reliability under background variation. This study addresses these limitations by introducing a two-stage training strategy and a three-condition evaluation protocol, providing quantitative evidence of background dependence through leaf-only and background-only testing. By formalizing background robustness using the 'Background Bias Coefficient' the proposed framework bridges the gap between laboratory accuracy and field reliability, contributing toward more trustworthy agricultural AI deployment.

## III. METHODOLOGY

### A. *Overview and System Architecture*

This study presents a two-stage training methodology designed to reduce background bias in plant disease detection while preserving high classification accuracy. This approach combines initial transfer learning and standard data augmentation to establish a baseline, followed by optimization training with background processing to encourage learning of disease-specific features rather than environmental shortcuts. Figure 1 gives the general architecture description of the projected plant disease detection system.

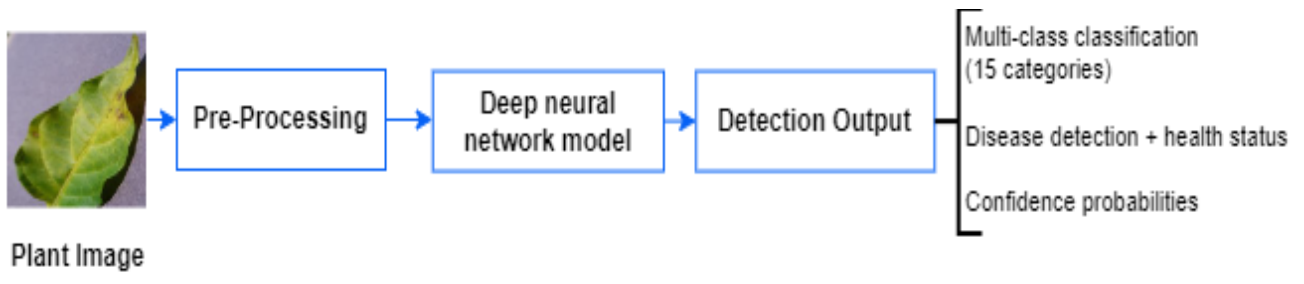

Fig. 1. High-level architecture of the proposed detection system.

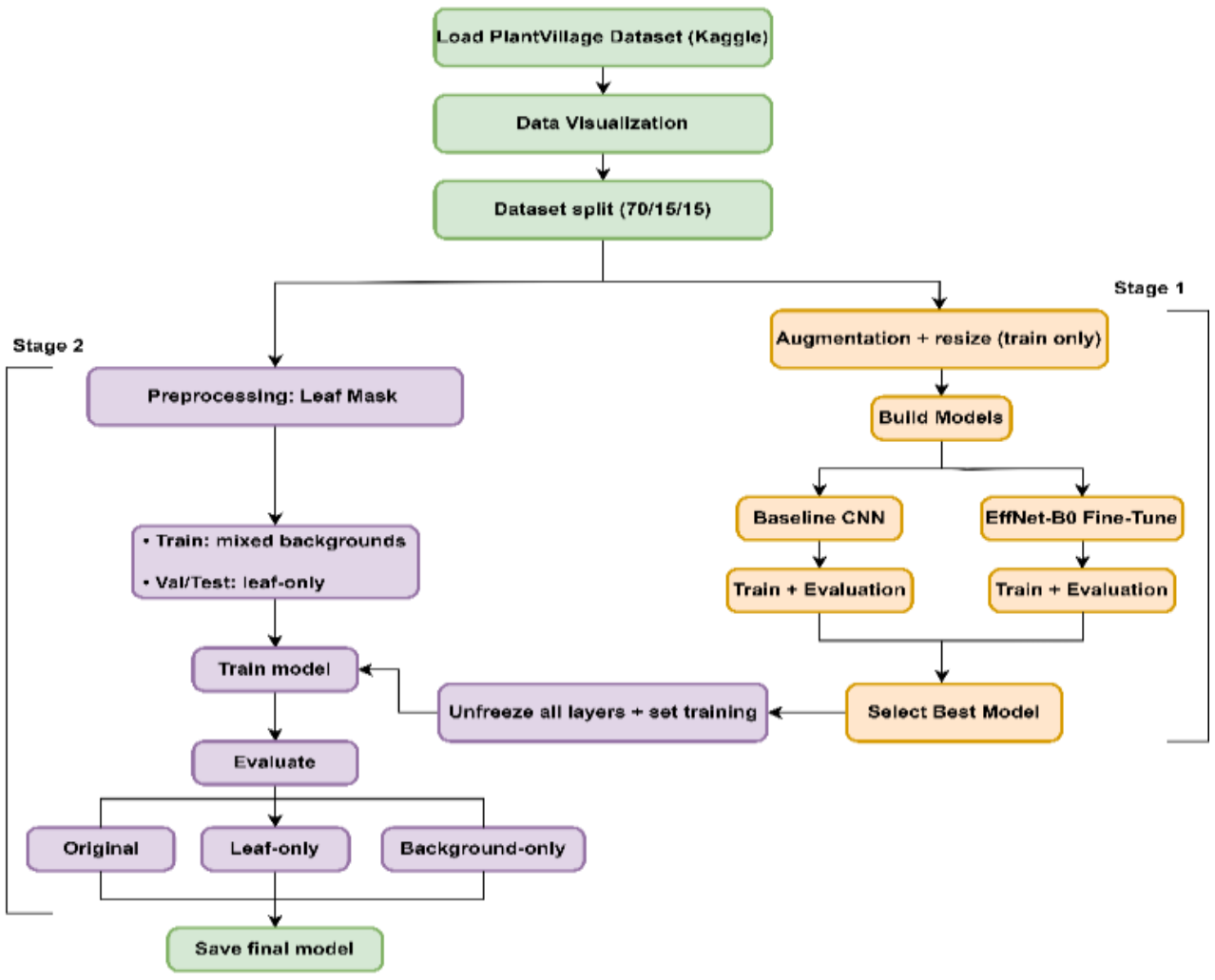

Fig. 2. Flowchart illustrating the two-stage end-to-end training pipeline

The complete training pipeline, illustrated in Figure 2 demonstrates the progression from Stage 1 baseline development through Stage 2 background bias elimination, incorporating specialized preprocessing functions and evaluation protocols.

## B. Dataset and Preprocessing

### 1) Dataset Description

The PlantVillage dataset was used as the primary data source, consisting of 20,638 high-resolution images spanning 15 disease classes across three crops: tomato, potato, and pepper. The dataset exhibits substantial class imbalance, with the largest class (Tomato Yellow Leaf Curl Virus) containing 3,209 images and the smallest class (Potato Healthy) only 152 images, corresponding to an imbalance ratio of approximately 21:1. Other major classes include Tomato Bacterial Spot (2,127 images) and Tomato Late Blight (1,909 images), while Tomato Mosaic Virus represents a mid-sized class with 373 images.

### 2) Data Splitting

A stratified data split was employed to preserve class distribution across subsets. The dataset was divided into training (70%), validation (15%), and test (15%) sets, ensuring adequate representation of minority classes in all partitions. The resulting split statistics are summarized in Table 1.

TABLE I. DATASET SPLIT STATISTICS

| | |
|---|---|
| **Training** | 14440 |
| **Validation** | 3089 |
| **Test** | 3109 |

### 3) Image Preprocessing Pipeline

All images were resized to 224×224 pixels to match the input requirements of EfficientNet-B0 [12]. During training, data augmentation was applied, including random rotations (±25°), width and height shifts (0.1), zoom (0.2), horizontal flipping, and brightness adjustment in the range [0.8, 1.2].

### 4) Background Processing Functions

To systematically control background information, a lightweight preprocessing pipeline was designed based on leaf segmentation using fixed HSV color thresholding combined with morphological operations. The segmentation mask $M(x, y)$ is defined using Equation 1: Mask generation function

$$M(x,y) = \begin{cases} 1, & \text{if } H(x,y) \in [H_{min}, H_{max}] \wedge S(x,y) \in [S_{min}, S_{max}] \wedge V(x,y) \in [V_{min}, V_{max}] \\ 0, & \text{otherwise} \end{cases}$$

where H, S, and V denote the hue, saturation, and value channels, with empirically selected thresholds.

This choice 'fixed HSV thresholding with morphological operations' was intentional to ensure simplicity, reproducibility, and to avoid introducing additional learned biases through a deep segmentation network. While this approach may be sensitive to lighting changes, non-green leaves, or severe discoloration, the same segmentation procedure is applied consistently across all experimental conditions. Therefore, relative comparisons—such as the Background Bias Coefficient—remain valid despite possible mask imperfections.

Based on this segmentation, three preprocessing variants were generated: (1) leaf-only images with isolated leaves on black backgrounds, (2) background-only images obtained by mask inversion, and (3) random-background images created by replacing original backgrounds with Gaussian-blurred noise. Representative examples are shown in Figure 3. Gaussian backgrounds were used as a simple stress test, while more realistic background substitutions are left for future work.

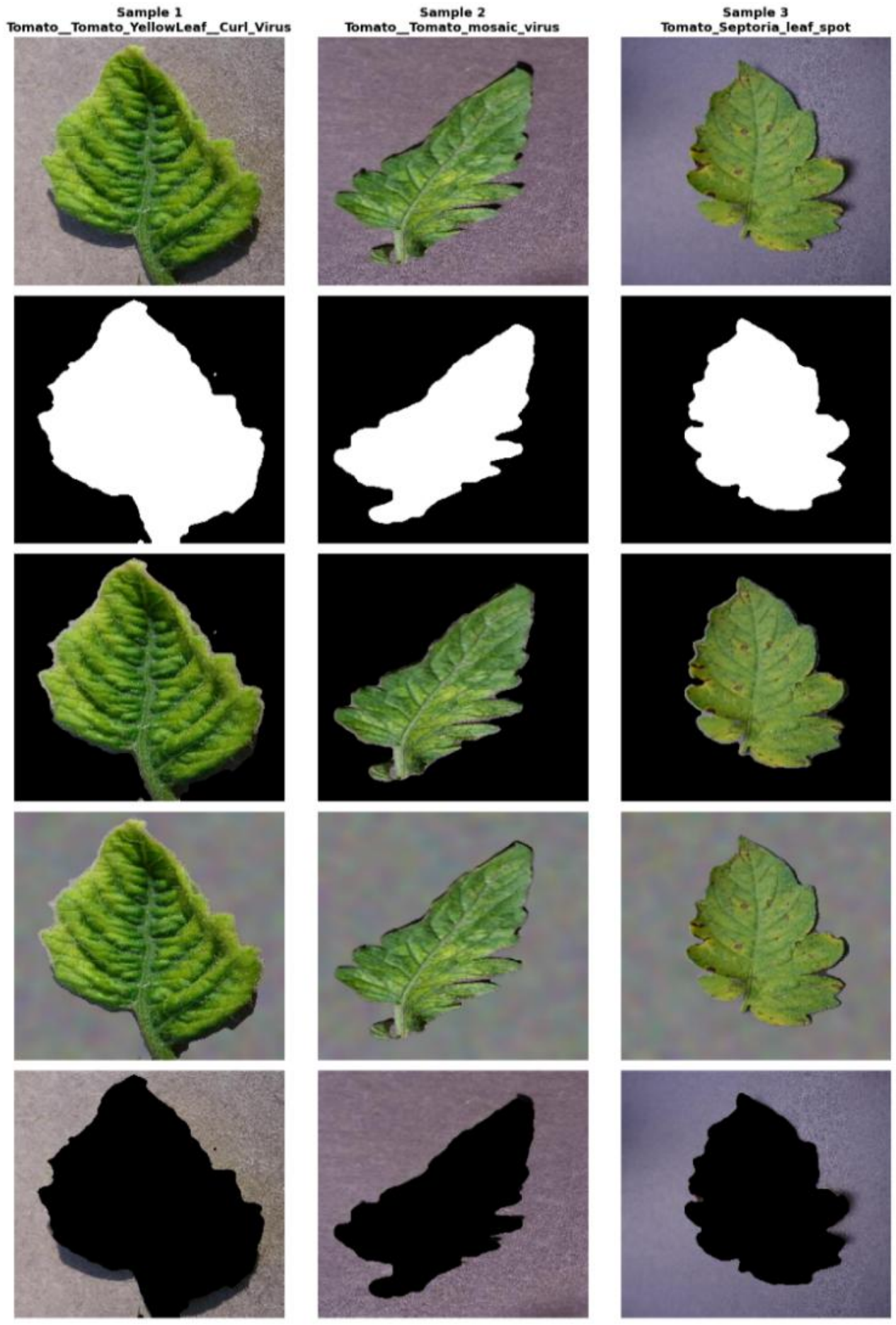

Fig. 3. Preprocessing examples: (a) original images, (b) leaf masks, (c) leaf-only, (d) random background replacement, and (e) background-only images.

## C. Two-Stage Training Framework

The proposed framework operates on a single EfficientNet-B0 model without multiple branches. Stage 1 establishes a baseline using conventional transfer learning on original images, while Stage 2 refines the same model through warm-start fine-tuning followed by mixed-background training using original, leaf-only, and randomized-background images.

### 1) Stage 1: Baseline Development

EfficientNet-B0 pre-trained on ImageNet was selected for its strong accuracy–efficiency trade-off [12]. A lightweight classification head (Global Average Pooling, Dense 256 + ReLU, Dropout 0.5, and a 15-class Softmax layer) was added. The last 40 backbone layers were unfrozen, while earlier layers remained frozen. We trained for 15 epochs using the Adam optimizer (lr = $3\times10^{-4}$) and early halting (patience = 4). A baseline CNN trained from scratch got 80.67% test accuracy, whereas the fine-tuned EfficientNet-B0 got 99.13%.

*2) Stage 2: Background Bias Elimination*

Stage 2 applies a two-phase refinement strategy. Phase A fine-tunes only the final layers for two epochs using AdamW (lr = $3\times10^{-4}$). Phase B unfreezes the last 40 layers and trains the model using mixed batches with equal proportions (33.3% each) of original, leaf-only, and randomized-background images (lr = $1\times10^{-4}$). This mixed-background training explicitly encourages robustness to background variations. Algorithm 1 summarizes the full two-stage procedure.

ALGORITHM 1: TWO-STAGE TRAINING PROCEDURE

**Input:**

D_original : Original PlantVillage dataset

D_leafonly: Leaf-only images (background removed)

D_randbg : Random background replacement images

N_classes : 15

**Output:**

M_final: Trained EfficientNet-B0 model robust to background Bias

**Stage 1: Baseline Training**

1: Initialize EfficientNet-B0 with ImageNet pre-trained weights

2: Add custom classification head:

GlobalAveragePooling→Dense(256, ReLU)→Dropout(0.5)

→ Dense(N_classes, Softmax)

3: Freeze early layers; unfreeze last 40 layers

4: Train on D_original for 15 epochs using:

Adam optimizer (lr = $3\times10^{-4}$) with cross-entropy loss, L2 regularization, and early stopping (patience = 4).

5: Save model weights as M_stage1

**Stage 2: Background Bias Elimination**

Phase A (Warm-start)

6: Load M_stage1

7: Freeze all except final 3 layers

8: Train for 2 epochs on D_original using:

Adam optimizer and class weights for imbalance

9: Save weights as M_A

Phase B (Refinement)

10: Unfreeze last 40 layers of M_A

11: Create mixed training batches:

33.3% D_original + 33.3% D_leafonly + 33.3% D_randbg

12: Train up to 8 epochs with:

Optimizer: AdamW (lr = $1\times10^{-4}$, wd = $1\times10^{-5}$)

Callbacks: ReduceLROnPlateau,

EarlyStopping(patience=4), ModelCheckpoint

13: Save best model weights as M_final

Return M_final

## *D. Evaluation Protocol*

*1) Three-Condition Testing*

To assess background dependency, the trained model was evaluated on a shared test set of 3,109 images under three conditions: (1) Original, using unmodified images; (2) Leaf-only, where backgrounds were removed to isolate disease-relevant features; and (3) Background-only, where leaf regions were masked to directly measure reliance on background information. Figure 4 illustrates the three evaluation conditions.

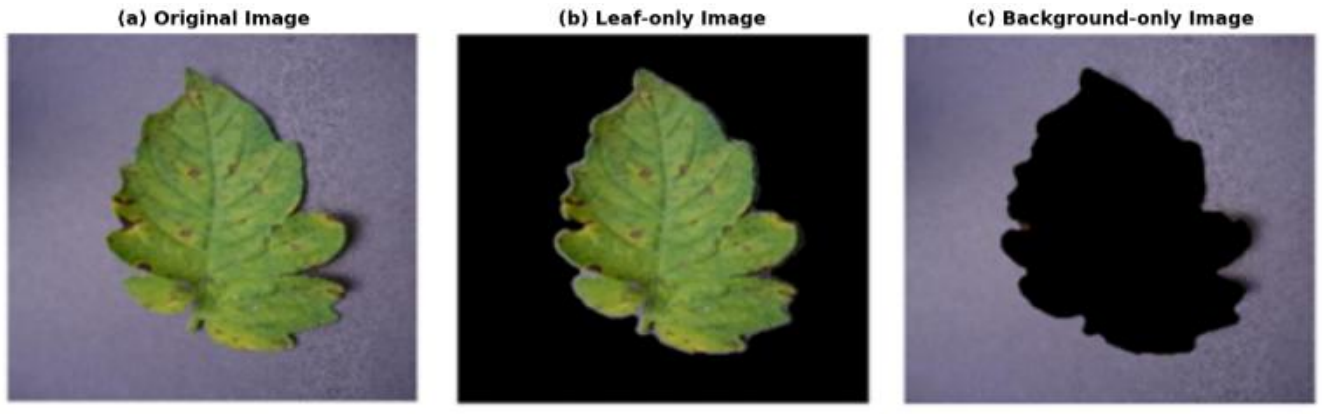

Fig. 4. Visual representation of the three-condition evaluation protocol: (a) original test image, (b) leaf-only test image, and (c) background-only test image

*2) Performance Metrics*

Model performance was assessed using accuracy, macro-averaged precision, recall, and Macro F1-score. Background dependency was quantified using the Background Bias Coefficient (BBC), defined using Equation 2:

$$BBC = (Acc_{original} - Acc_{background}) \,/\, Acc_{original}$$

where values closer to 1 indicate lower background dependence.

# IV. RESULTS AND DISCUSSION

## *A. Training Performance*

*1) Model Comparison*

Comparative evaluation confirms the efficacy of transfer learning. The baseline CNN trained from scratch achieved 80.67% correct on the test (Macro-F1 = 0.749), but the fine-tuned EfficientNet-B0 achieved 99.13% correct (Macro-F1 = 0.990), which is an improvement of 18.46 percentage points (Table II). Figure 5 shows that EfficientNet-B0 converges faster and has more stable optimization than the baseline CNN.

TABLE II. MODEL PERFORMANCE COMPARISON

| Model | Accuracy | Precision | Recall | Macro F1 |
|---|---|---|---|---|
| Baseline CNN | 0.8153 | 0.7605 | 0.7512 | 0.7489 |
| Fine-tuned EfficientNet-B0 | 0.9958 | 0.9962 | 0.9949 | 0.9955 |

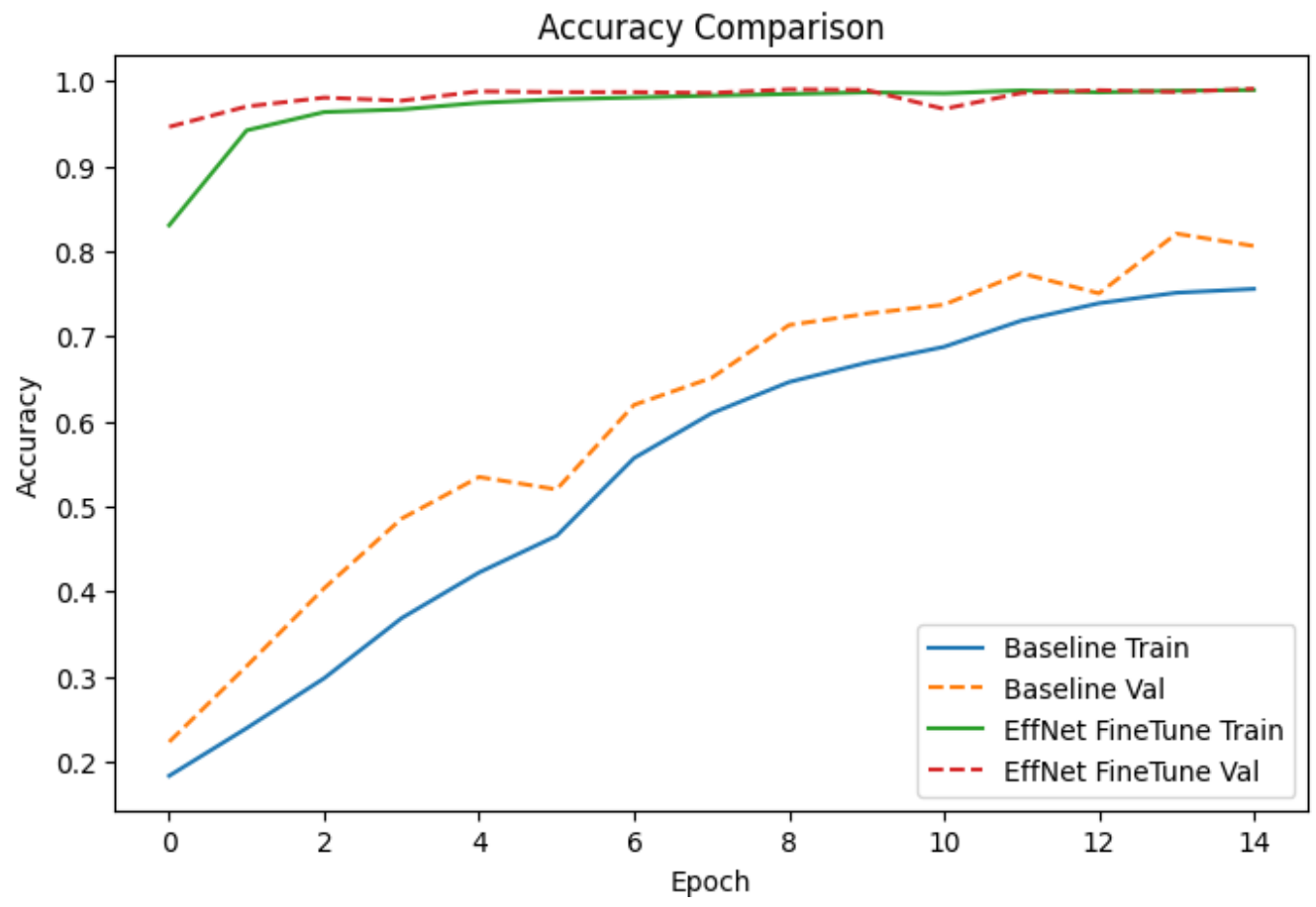

Fig. 5. Training/validation accuracy curves comparison

*2) Stage 2 Training Dynamics*

Stage 2 training improved background robustness while preserving accuracy. Validation accuracy increased from 91.74% to 93.85% during the warm-start phase and reached 97.05% during refinement before early stopping. The final model achieved 98.97% test accuracy, reflecting a minor drop from Stage 1 (99.13%) that is acceptable given the substantial gain in background invariance.

## B. Three-Condition Evaluation Results

Systematic evaluation under three test conditions provides quantitative evidence of background bias reduction. Table III summarizes the model performance under the three evaluation conditions, also visualized using a bar chart in Figure 6.

TABLE III. PERFORMANCE UNDER THREE EVALUATION CONDITIONS

| Evaluation Setting | Accuracy (%) | Macro-F1 |
|---|---|---|
| Original images | 98.97 | 0.9901 |
| Leaf-only images | 96.62 | 0.9635 |
| Background-only | 23.58 | 0.2590 |

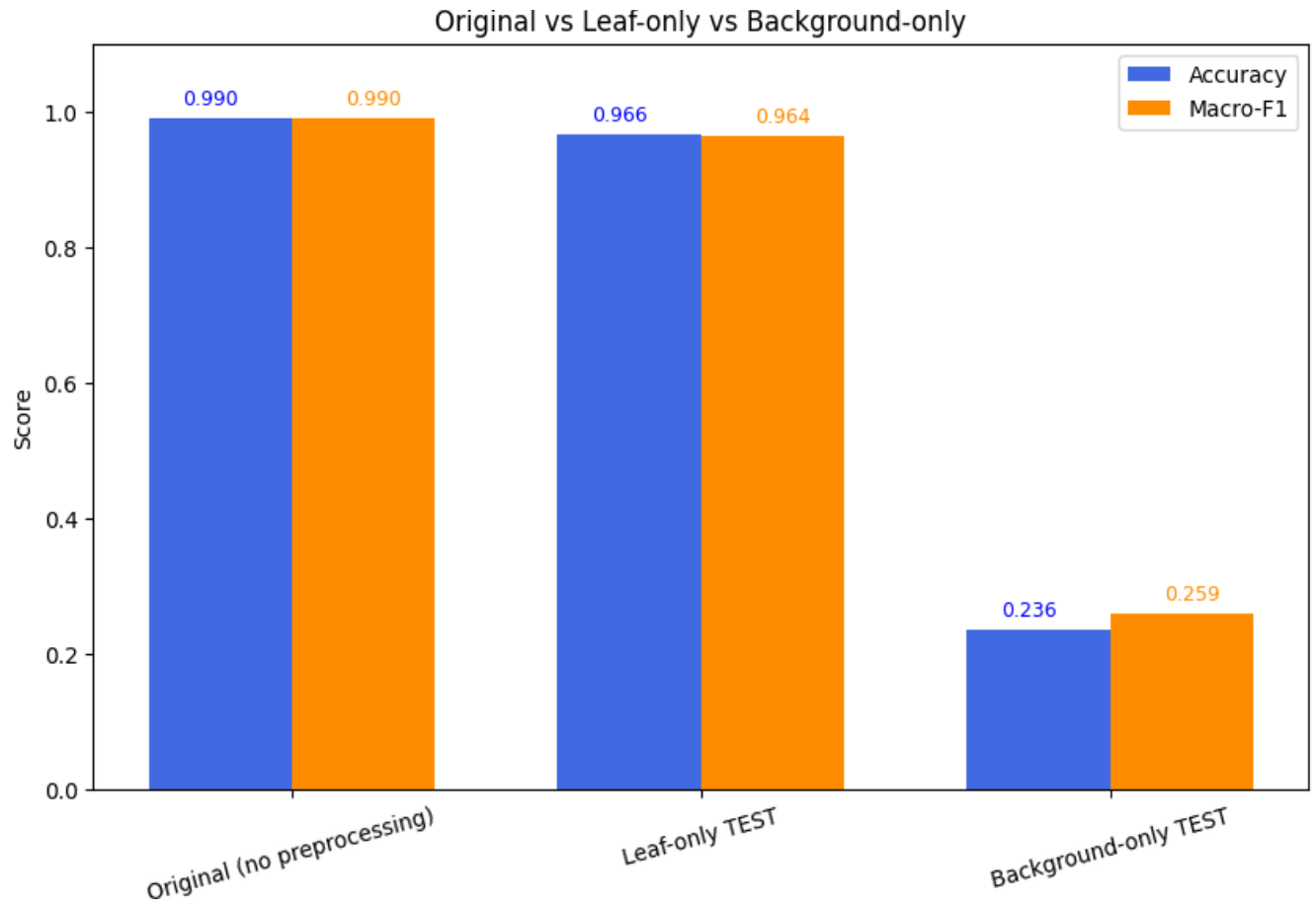

Fig. 6. Accuracy and Macro-F1 across three test conditions

The model achieves 98.97% accuracy on original images and 96.62% accuracy on leaf-only images, which means that the predicted performance is mostly based on leaf features related to disease. In contrast, accuracy drops to 23.58% when only the background is present, which remains above random chance (6.7% for 15 classes) but far below meaningful classification performance. This confirms that background information alone is insufficient for reliable disease recognition.

The residual background-only performance is partially attributed to imperfect leaf masking, where geometric leaf silhouettes, boundaries, or spatial cues may persist after color removal (Fig. 4(c)), as well as to the largely uniform and laboratory-controlled acquisition protocol of the PlantVillage dataset. The resulting Background Bias Coefficient (BBC) of 0.762 indicates that approximately 76.2% of the model's predictive capability is attributable to leaf-related visual features rather than background correlations. While BBC serves as a global summary metric, future work may explore class-weighted variants to capture class-specific sensitivities.

## C. Confusion Matrix Analysis

Figure 7 illustrates the confusion matrix, which indicates that most classes are almost perfectly classified. Misclassifications largely arise between visually analogous tomato illnesses, such as Early Blight and Target Spot, indicating that symptom overlap, rather than background bias, is the principal source of residual mistakes.

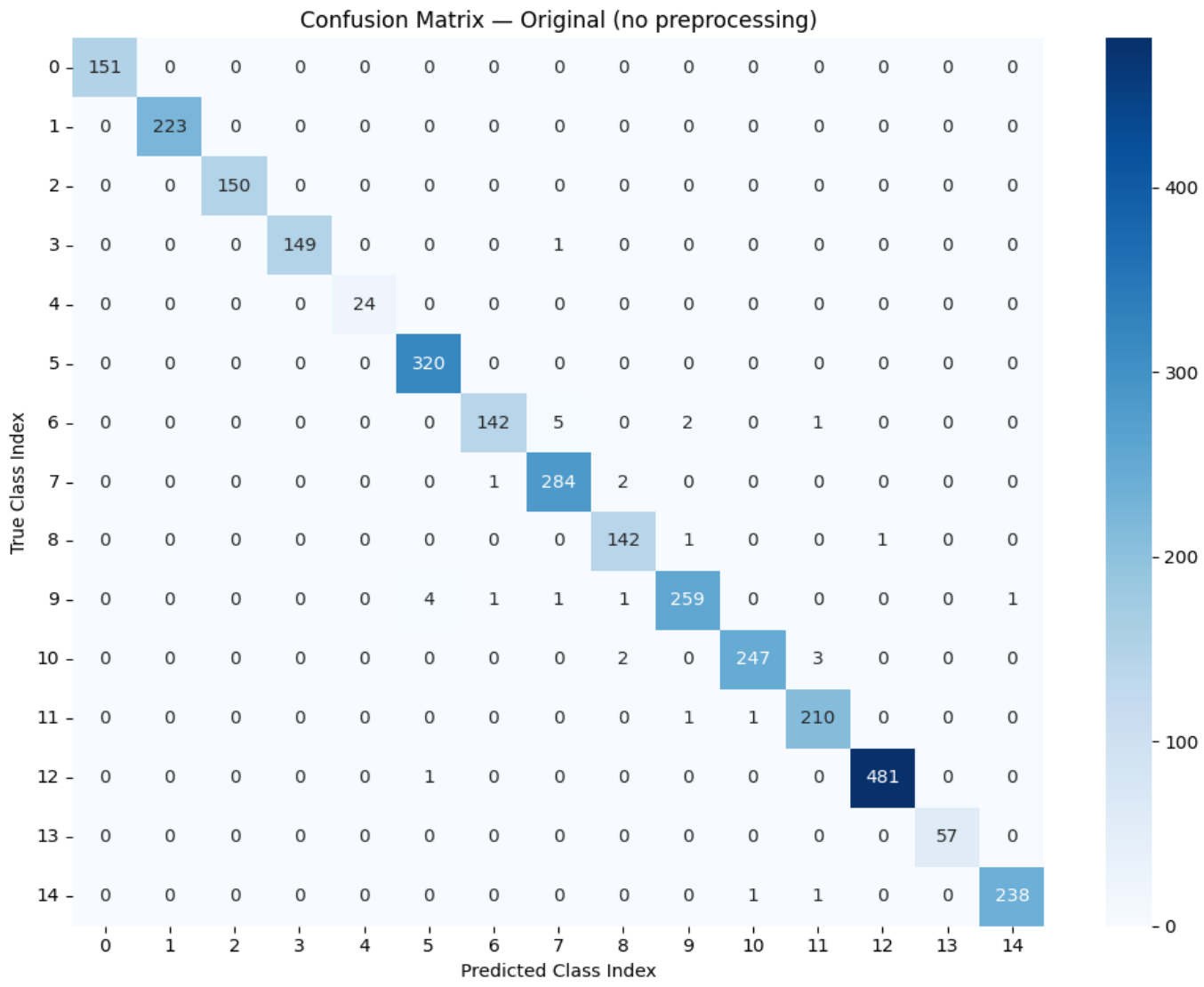

Fig. 7. Confusion matrix of the proposed model under the original test

## D. Qualitative Analysis

A qualitative examination of model predictions under various test situations elucidates the model's decision-making processes. Representative samples (Figure 8) show that predictions of high confidence are consistently accurate across different illness classes. This shows that the model can generalize well even when the background noise is low. The analysis of misclassified samples shows systematic problems, mostly because it's hard to tell the difference between early and late disease stages, there is modest mechanical damage on healthy leaves.

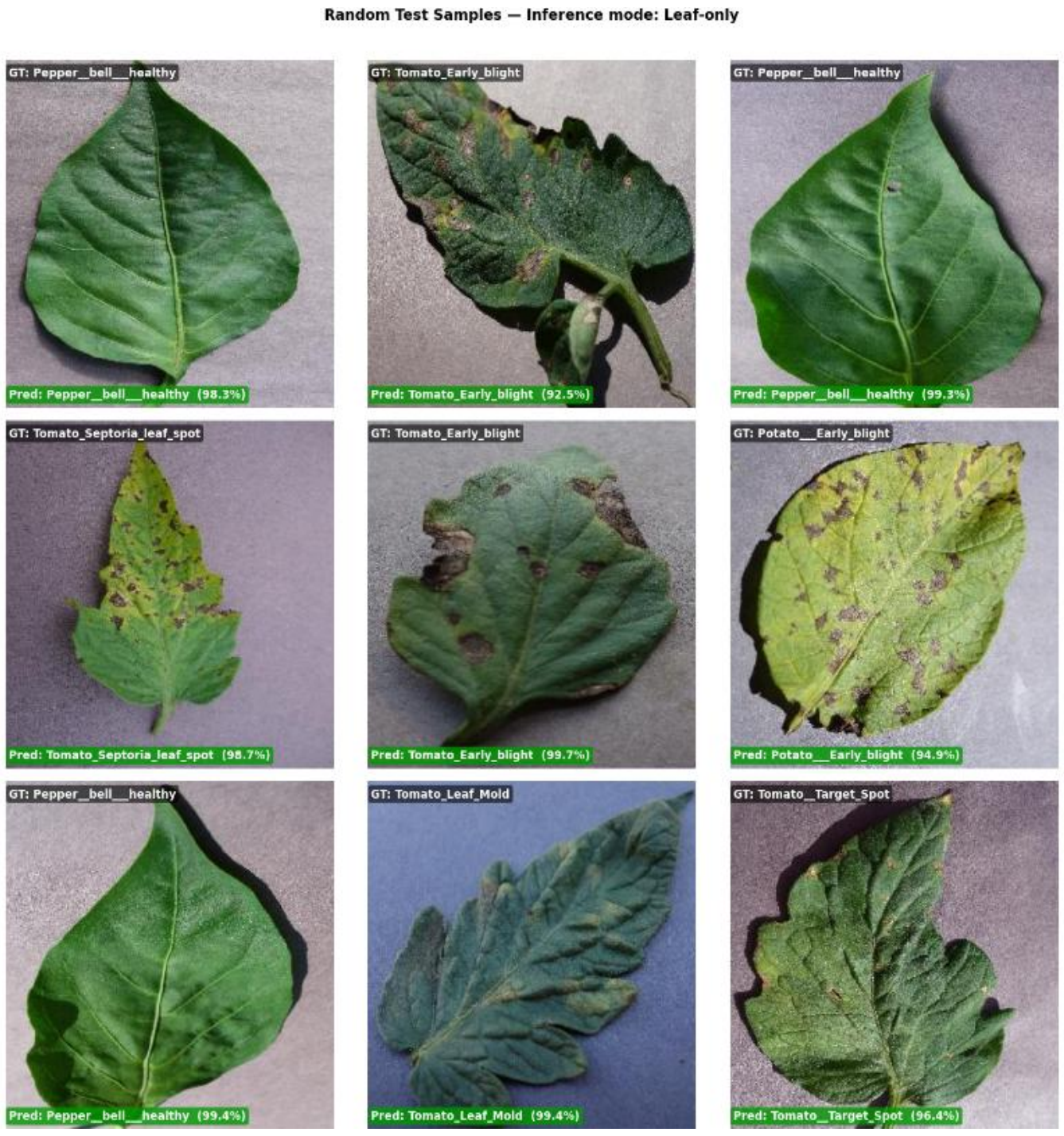

Fig. 8. Grid of test samples with predictions

## V. CONCLUSION

This study presents a systematic and quantitative investigation of background bias in plant disease detection, exposing a critical limitation of high-accuracy deep learning models trained on controlled datasets. Through a two-stage training framework and a three-condition evaluation protocol, we show that even high-performing models may partially rely on environmental shortcuts rather than exclusively on pathological features. Although EfficientNet-B0 achieves strong performance on original and leaf-only images, the substantial drop in background-only accuracy confirms that background information alone cannot support reliable disease classification. The proposed Background Bias Coefficient (BBC) further quantifies this effect, indicating that more than 76% of predictive performance is driven by leaf-related visual features.

Training only on segmented images can help reduce background bias; however, it may also limit robustness when deployed in unconstrained field environments. The proposed mixed-background training strategy offers a balanced alternative by preserving contextual awareness while explicitly penalizing reliance on spurious background correlations.

Further studies will extend this evaluation approach to encompass field-acquired datasets, investigate alternative segmentation and attention-based methods, and evaluate the impact of other model designs, including Vision Transformers. This work together develops a pragmatic bias-evaluation and mitigation paradigm that propels plant disease detection towards dependable real-world agricultural implementation.

## VI. CODE AVAILABILITY

The complete implementation, including preprocessing, training, and evaluation scripts, is publicly available to ensure reproducibility [13].

## REFERENCES

[1] S. Savary, L. Willocquet, S. J. Pethybridge, P. Esker, N. McRoberts, and A. Nelson, "The global burden of pathogens and pests on major food crops," *Nat. Ecol. Evol.*, vol. 3, no. 3, pp. 430–439, Feb. 2019, doi:10.1038/s41559-018-0793-y.

[2] I. Pacal *et al.*, "A systematic review of deep learning techniques for plant diseases," *Artif. Intell. Rev.*, vol. 57, no. 11, Nov. 2024, doi: 10.1007/s10462-024-10944-7.

[3] S. P. Mohanty, D. P. Hughes, and M. Salathé, "Using Deep Learning for Image-Based Plant Disease Detection," *Front. Plant Sci.*, vol. 7, Sep. 2016, doi: 10.3389/fpls.2016.01419.

[4] A. Lakshmanarao, N. Supriya, and A. Arulmurugan, "Plant Disease Prediction using Transfer Learning Techniques," in *2022 2nd International Conference on Advances in Electrical, Computing, Communication and Sustainable Technologies, ICAECT 2022*, Institute of Electrical and Electronics Engineers Inc., 2022. doi: 10.1109/ICAECT54875.2022.9807956.

[5] T. H. Nguyen, T. N. Nguyen, and B. V. Ngo, "A VGG-19 Model with Transfer Learning and Image Segmentation for Classification of Tomato Leaf Disease," *AgriEngineering*, vol. 4, no. 4, pp. 871–887, Dec. 2022, doi: 10.3390/agriengineering4040056.

[6] H. Ali, N. Shifa, R. Benlamri, A. A. Farooque, and R. Yaqub, "A fine tuned EfficientNet-B0 convolutional neural network for accurate and efficient classification of apple leaf diseases," *Sci. Rep.*, vol. 15, no. 1, Dec. 2025, doi: 10.1038/s41598-025-04479-2.

[7] M. Seshaiah, D. K. Shravya, K. H. Tarun, R. Vandana, and R. Varshitha, "Recognition of Plant disease using Deep Learning Approaches," in *2024 International Conference on Knowledge Engineering and Communication Systems, ICKECS 2024*, Institute of Electrical and Electronics Engineers Inc., 2024. doi: 10.1109/ICKECS61492.2024.10616775.

[8] S. Karthikeyan, R. Charan, S. Narayanan, and L. Jani Anbarasi, "Enhanced plant disease classification with attention-based convolutional neural network using squeeze and excitation mechanism," *Front. Artif. Intell.*, vol. 8, 2025, doi: 10.3389/frai.2025.1640549.

[9] M. U. I. Tamim, S. A. Hamim, S. Malik, M. F. Mridha, and S. Mahmood, "InsightNet: A Deep Learning Framework for Enhanced Plant Disease Detection and Explainable Insights," *Plant Direct*, vol. 9, no. 5, May 2025, doi: 10.1002/pld3.70076.

[10] R. Geirhos *et al.*, "Shortcut learning in deep neural networks," *Nat. Mach. Intell.*, vol. 2, no. 11, pp. 665–673, Nov. 2020, doi: 10.1038/s42256-020-00257-z.

[11] S. Lapuschkin, S. Wäldchen, A. Binder, G. Montavon, W. Samek, and K.-R. Müller, "Unmasking Clever Hans predictors and assessing what machines really learn," *Nat. Commun.*, vol. 10, no. 1, p. 1096, Mar. 2019, doi: 10.1038/s41467-019-08987-4.

[12] M. Tan and Q. V. Le, "EfficientNetV2: Smaller Models and Faster Training," Jun. 2021, [Online]. Available: http://arxiv.org/abs/2104.00298

[13] M. Molhi, "majedmolhi/plant-disease-background-bias: Initial Release (Plant Disease Detection with Background Bias Elimination)". Zenodo, Jan. 15, 2026. doi: 10.5281/zenodo.18258119.

[14] M. S. Fathima and B. Booba, "Enhancing Plant Leaf Disease Detection: Integrating Mobilenet with Local Binary Pattern and Visualizing Insights with Grad-Cam," in *Proceedings of International Conference on Circuit Power and Computing Technologies, ICCPCT 2024*, Institute of Electrical and Electronics Engineers Inc., 2024, pp. 1400–1404. doi: 10.1109/ICCPCT61902.2024.10673312.

[15] R. S. Joseph, Nisha, and K. K. Shandry, "Plant Disease Detection: A Survey on Recent Deep Learning Technologies," in *Proceedings - 2nd International Conference on Next Generation Intelligent Systems, ICNGIS 2022*, Institute of Electrical and Electronics Engineers Inc., 2022. doi: 10.1109/ICNGIS54955.2022.10079766.

[16] A. Gehlawat, "A comparative analysis of different CNN models for plant disease detection and classification," *International Journal of Advanced Research in Computer and Communication Engineering Impact Factor*, vol. 8, 2024, doi: 10.17148/IJARCCE.2024.13616.

[17] N. Lokhande, V. Thool, and P. Vikhe, "Comparative analysis of different plant leaf disease classification and detection using CNN," in *2024 International Conference on Recent Innovation in Smart and Sustainable Technology, ICRISST 2024*, Institute of Electrical and Electronics Engineers Inc., 2024. doi: 10.1109/ICRISST59181.2024.10921975.